# Local Quadruple Pattern: A Novel Descriptor for Facial Image Recognition and Retrieval

[1]Soumendu Chakraborty, [1]Satish Kumar Singh, and [1]Pavan Chakraborty

[1]Indian Institute of Information Technology, Allahabad, India
soum.uit@gmail.com, sk.singh@iiita.ac.in, pavan@iiita.ac.in

**ABSTRACT**

In this paper a novel hand crafted local quadruple pattern (LQPAT) is proposed for facial image recognition and retrieval. Most of the existing hand-crafted descriptors encodes only a limited number of pixels in the local neighbourhood. Under unconstrained environment the performance of these descriptors tends to degrade drastically. The major problem in increasing the local neighbourhood is that, it also increases the feature length of the descriptor. The proposed descriptor try to overcome these problems by defining an efficient encoding structure with optimal feature length. The proposed descriptor encodes relations amongst the neighbours in quadruple space. Two micro patterns are computed from the local relationships to form the descriptor. The retrieval and recognition accuracies of the proposed descriptor has been compared with state of the art hand crafted descriptors on bench mark databases namely; Caltech-face, LFW, Colour-FERET, and CASIA-face-v5. Result analysis shows that the proposed descriptor performs well under uncontrolled variations in pose, illumination, background and expressions.

## 1. Introduction

The feature description using hand-crafted descriptors has gained more attention in the recent past due to its capability to encode individual images without any supervised learning. Such descriptors have been proposed to encode more pixels in the local neighbourhood to enhance recognition and retrieval accuracies under varying pose, illumination, expression and background. The fundamental requirement of any descriptor is that, it should increase the inter class dissimilarity and decrease the intra class dissimilarity. Primary objective of the descriptors such as; Eigen-face [1], Fisher-face [1], variations of PCA [2][3][4][5][6], and Linear Discriminant Analysis (LDA) [7][8][9] is to identify the feature points of the facial images taken under constrained environment. Two dimensional and Two Directional Random Projection has been used in [10] to reduce the dimensionality of the feature matrix of an image. The major advantage of this method is that, it removes the problem of singularity, SSS (small sample size) and over-fitting [10]. DWDPA [11-12] is a unique feature description technique, which increases the discrimination power by weighting the dominant selected DCT coefficients [11-12]. Similar strategy can be adopted with spatial domain image description by computing the histogram and determining the dominant features and weighting them to improve the discriminating power.

Convolutional Neural Network (CNN) has been used to extract features from facial images by training the CNN with relevant datasets [13-14]. These learning based descriptors require large training datasets to achieve optimal recognition accuracy. These descriptors are generally application specific due to the biasness towards the training databases [15-16]. The proposed descriptor belongs to the class of hand-crafted descriptors, which captures the core information from the inherent pixel relations. It does not require any learning (training) mechanism to represent images (single or multiple) of a given class. Apart from this, the characteristic of the proposed descriptor is very much different from VGG face recognition model that make use of Convolutional Neural Networks (CNNs) to extract the features from an image.

One of the earliest descriptor defining the relationships in the local neighbourhood is Local Binary Pattern (LBP) [17-18]. Eight neighbours of the reference pixels are encoded to generate the binary pattern. Centre Symmetric Local Binary Pattern (CSLBP) [19] has been derived from LBP to achieve improvements in region based image matching and length. Illumination variation is another fundamental problem in face recognition addressed through Centre Symmetric Local Ternary Pattern (CSLTP) [20]. It is a gradient based local descriptor, which works well in controlled illumination variation. Most recently Multi-Block Local Binary Pattern (MB-LBP) has been used detect pedestrians [21]. Region wise averages used to define local descriptor such as Semi Local Binary Pattern (SLBP) [22], to improve the recognition and retrieval accuracies under scale, noise and illumination variations.

There is another class of descriptors defined in the higher order derivative space. Local Directional Gradient Pattern (LDGP) [23] is one of the most recent higher order descriptor, which shows improvements over Local Derivative Pattern (LDP) [24], and Local Vector Pattern (LVP) [25] with respect to time and achieves comparable recognition rates.

The proposed descriptor is a region based descriptor, which captures distinctive information in the larger region with optimal feature length and improved recognition and retrieval accuracies.

The organization of the rest of the paper is as follows. Some of the similar descriptors are elaborated in section 2. Section 3 elaborates the motivation and proposition of the descriptor. Various experiments have been performed and the obtained results are compared with the state of the art descriptors in section 4. The work reported through this paper is concluded in section 5.

## 2. Overview of similar descriptors

In this section we give a small overview of the most closely related descriptors.

*2.1 LBP*

One of the most commonly used descriptors; local binary pattern (LBP) captures the relationship amongst the pixels in the local neighbourhood of the reference pixel. Fig.1. shows the template of the operator. Centre pixel $\mathbb{R}_0$ is the reference pixel and remaining eight pixels $\mathbb{R}_1..\mathbb{R}_8$ are the neighboring pixels surrounding the reference pixel.

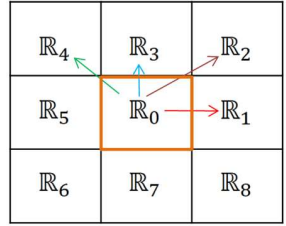

Fig. 1. Template of the LBP operator.

LBP is defined as a binary string generated from the encoding function $B(.)$.

$$\text{LBP}(\mathbb{R}_0) = \{B(I(\mathbb{R}_0), I(\mathbb{R}_1)), B(I(\mathbb{R}_0), I(\mathbb{R}_2))..B(I(\mathbb{R}_0), I(\mathbb{R}_n))\}|_{n=8} \quad (1)$$

where $\mathbb{R}_0$ is the reference point (pixel) in the image and $I(\mathbb{R}_0)$ returns the intensity of the pixel. $B(.)$ is the encoding function defined as

$$B(I(\mathbb{R}_0), I(\mathbb{R}_n)) = \begin{cases} 1 & \text{if } I(\mathbb{R}_n) - I(\mathbb{R}_0) > T \\ 0 & \text{if } I(\mathbb{R}_n) - I(\mathbb{R}_0) \leq T \end{cases} \quad (2)$$

where T is the threshold.

Although the LBP is a rotation invariant descriptor, but fails while tested on complex facial datasets (i.e. sever changes in illumination, pose, light, scale, and expression etc.).

*2.2 CSLBP*

The structure of CSLBP is such that it requires only 4 bits to encode the local neighbourhood. The length of this local descriptor is only 4 bits long and encodes only 8 pixels. The CSLBP is defined using the template shown in Fig.1. as follows

$$CSLBP(\mathbb{R}_0) = \{B(I(\mathbb{R}_1), I(\mathbb{R}_5)), B(I(\mathbb{R}_2), I(\mathbb{R}_6)), B(I(\mathbb{R}_3), I(\mathbb{R}_7)), B(I(\mathbb{R}_4), I(\mathbb{R}_8))\} \quad (3)$$

CSLBP generates the binary code using the same encoding function defined in (2).

## 3. Proposed local quadruple pattern

*3.1 Motivation*

The need of the proposed descriptor with increased neighborhood and low dimension is summarized as follows

1) Distinctive information that exists in a facial image tends to spread around the local region of the reference pixel. Most of the descriptors confine the local region to the pixels on the circumference of a circle ignoring the pixels within the circular region. If we try to capture the relationship of the reference pixel with the pixels within the circular neighborhood then the length of the micropattern increases beyond the acceptable limits.

2) Motivated from the existing local descriptors like CSLBP, CSLTP, LDP, LVP, LBP, SLBP and so forth, we propose a descriptor which not only captures additional distinctive information that exists in the local region but also maintains the length within the acceptable limits.

3) The proposed descriptor encodes 15 pixels in local neighborhood of the reference pixel. The information contained by the feature images generated by different descriptors is numerically measured using entropy.

4) The entropy of the feature images are computed from the probability distribution of the grayscales of the feature images. As LDP and LVP generates four feature images, the entropy of LDP and LVP are computed as the average of entropy of four feature images. Similarly entropy of LQPAT is the average of the entropy of two feature images.

5) The average entropies of proposed and state of the art descriptors are computed for the feature images generated over Caltech-Face [18] database.

6) The average entropy of CSLBP, CSLTP, LDP, LVP, LBP, SLBP, and LQPAT are 2.75, 1.77, 5.97, 6.67, 3.87, 3.60 and 7.21.

7) Result analysis confer that these additional distinctive relationships are useful enough to significantly increase the accuracy of the proposed descriptor under constrained and unconstrained environment.

*3.2 Major contribution*

The proposed descriptor extends the local neighborhood to $4 \times 4$ block and constructs the micropattern in such a way that the feature length of the descriptor does not increase beyond acceptable limits. The major contribution of the proposed descriptor is summarized as follows

1) LBP encodes only eight neighbors in the local neighborhood of the reference pixel and generates an eight bit pattern. The proposed descriptor encodes 16 pixels including the reference pixel in the local neighborhood with eight bits.

2) SLBP takes sum of the pixels in $2 \times 2$ blocks and then computes the LBP over these summations. SLBP loses significant discriminating information while summing the pixel values.

3) Although CSLBP and CSLTP reduces the feature length of the descriptor, they only consider four neighbors in the local neighborhood of the reference pixel.

4) Most of the descriptors namely; LDP, LVP, LDGP and so forth proposed in higher order derivative space consider only eight neighbors in the construction of the micropattern. On the other hand proposed Local Quadruple Pattern (LQPAT) encodes fifteen pixels in the local neighborhood of the reference pixel.

5) LQPAT computes the micropattern of length 16 bits which is notably less compared to the 32 bit patterns computed by LDP and LVP. Feature lengths shown in Table 1 presents a clear view on how LQPAT efficiently encodes more neighbors without increasing the feature length too much.

*3.3 Proposed descriptor*

Local pattern descriptors try to capture local relationships amongst the pixels and encode it into binary codes which effectively recognize the facial image under illumination, pose and expression variations. Proposed Local Quadruple

Pattern (LQPAT) captures relationship amongst the pixels across four square blocks of size $2 \times 2$.

Fig. 2. shows the template of the local neighborhood of the reference pixel $I_{i,j}$ of image $I$ of size $M \times N$. The reference pixel is the current pixel being encoded. The template is a sliding window, which moves over the entire image. The local neighborhood of the reference pixel is divided into four (quadruple) blocks shown with red, green, blue and purple colors. Pixels of red block are compared with the respective pixels of green block to compute the first micropattern and the corresponding decimal equivalent using (4).

Similarly, Pixels of green block are compared with the respective pixels of blue block to compute the second micropattern and the corresponding decimal equivalent using (5). The micropatterns and corresponding decimal equivalents for blue and purple block pair and purple and red block pair are computed using (6) and (7) respectively.

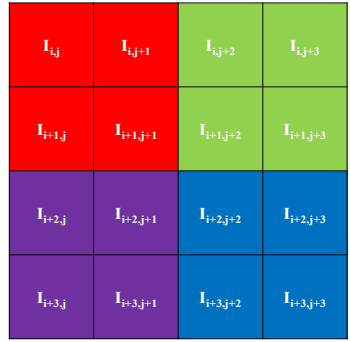

Fig. 2. Template showing the local neighborhood of the reference pixel $I_{i,j}$.

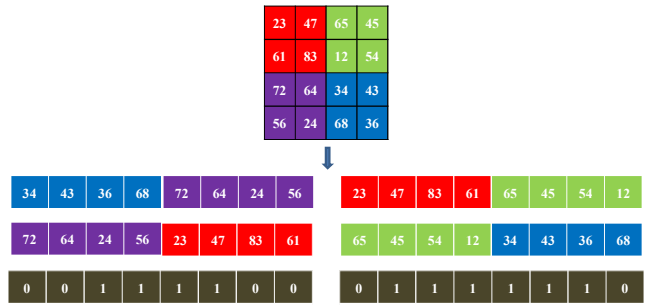

Fig. 3. Example showing the encoding scheme of LQPAT.

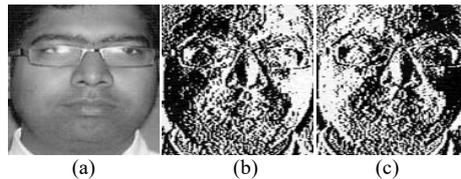

(a)      (b)      (c)

Fig. 4. (a) Original image, (b) Feature image computed using red, green and blue blocks, (c) Feature image computed using blue, purple, and red blocks.

$$A_{i,j}^1 = 2^7 \times C(I_{i,j}, I_{i,j+2}) + 2^6 \times C(I_{i,j+1}, I_{i,j+3}) + 2^5 \times C(I_{i+1,j}, I_{i+1,j+2}) + 2^4 \times C(I_{i+1,j+1}, I_{i+1,j+3}) \quad (4)$$

$$A_{i,j}^2 = 2^3 \times C(I_{i,j+2}, I_{i+2,j+2}) + 2^2 \times C(I_{i,j+3}, I_{i+2,j+3}) + 2^1 \times C(I_{i+1,j+2}, I_{i+3,j+2}) + 2^0 \times C(I_{i+1,j+3}, I_{i+3,j+3}) \quad (5)$$

$$B_{i,j}^1 = 2^7 \times C(I_{i+2,j+2}, I_{i+2,j}) + 2^6 \times C(I_{i+2,j+3}, I_{i+2,j+1}) + 2^5 \times C(I_{i+3,j+2}, I_{i+3,j}) + 2^4 \times C(I_{i+3,j+3}, I_{i+3,j+1}) \quad (6)$$

$$B_{i,j}^2 = 2^3 \times C(I_{i+2,j}, I_{i,j}) + 2^2 \times C(I_{i+2,j+1}, I_{i,j+1}) + 2^1 \times C(I_{i+3,j}, I_{i+1,j}) + 2^0 \times C(I_{i+3,j+1}, I_{i+1,j+1}) \quad (7)$$

Where $i = 1,2..M-3$, $j = 1,2...N-3$ and $C$ is the encoding function defined as

$$C(E,F) = \begin{cases} 0, & if E \leq F \\ 1, & else \end{cases} \quad (8)$$

Published in "Computers & Electrical Engineering, vol-62, pp. 92-104, (2017). (Elsevier) ISSN/ISBN: 0045-7906"

$A_{i,j}^1$ and $A_{i,j}^2$ are combined to compute the decimal equivalent of the 8 bit pattern generated from red, green, and blue blocks in (9).

$$A = A_{i,j}^1 + A_{i,j}^2 \qquad (9)$$

Similarly, decimal equivalent $B$ of the 8 bit pattern generated from blue, purple, and red blocks is computed in (10).

$$B = B_{i,j}^1 + B_{i,j}^2 \qquad (10)$$

Finally histograms of $A$ and $B$ denoted as $H_A$ and $H_B$ respectively are concatenated to form the $LQPAT$ feature vector as shown in (11).

$$LQPAT = \{H_A, H_B\} \qquad (11)$$

$\chi^2$ distance [26] is used to measure the similarity between two histograms. Similarity measure $S_{\chi^2}(.,.)$ is defined as

$$S_{\chi^2}(X,Y) = \frac{1}{2}\sum_{i=0}^{q} \frac{(x_i - y_i)^2}{(x_i + y_i)} \qquad (12)$$

where $S_{\chi^2}(X,Y)$ is the $\chi^2$ distance computed on two vectors $X = (x_1, ..., x_q)$ and $Y = (y_1, ..., y_q)$. Similarity of the probe and gallery features are computed using $\chi^2$ distance Class of the probe images are identified using one nearest neighbor (1NN) classifier [24]. 1NN is used for the classification to reduce the computation cost of the overall system[24].

A sample image and the encoding structure has been depicted in Fig.3. Encoding function (8) has been used to generate the 8 bit pattern by comparing the pixel intensities in red and green blocks with the pixel intensities in green and blue blocks respectively. The resulting 8 bit pattern has been shown in gray on the right side of Fig. 3. Similarly, another 8 bit pattern has been computed by comparing the pixel intensities in blue and purple blocks with the pixel intensities in purple and red blocks respectively. The resulting 8 bit pattern has been shown in gray on the left side of Fig. 3. Finally these two binary patterns are converted to equivalent decimal values to generate two feature images.

These feature images are shown in Fig. 4(b) and (c) for the original image shown in Fig. 4(a). Feature images clearly show that the descriptor captures distinctive complementary relationships that exist in the local neighborhood of the reference pixel.

**4 performance analysis**

Performance of the proposed $LQPAT$ has been analyzed with respect to retrieval accuracies. Performance measures used to evaluate retrieval accuracy are Average Retrieval Precision (ARP) and Average Retrieval Rate (ARR). Each image is used as the query image to retrieve varying number of top matched images. Average of precision [27] and average of recall [27] are computed over each class with each image taken as a query. ARP and ARR are computed by taking the means of the average precision and average recall respectively.

Precision is defined as the number of relevant images retrieved out of the total number of retrieved images and Recall is defined as the total number of relevant images retrieved. As the denominator of the recall rate is the total number of images per class, which is constant for a particular class. Hence, with increasing number of retrieved images, recall rate should also increase.

Average normalized modified retrieval rank (ANMRR) defined in [27-28] is used to measure the performance of the descriptors based on the rank of the retrieved images. Low ANMRR indicates that the images retrieved by the descriptor are highly relevant to the queried image and higher value of ANMRR indicates that most of the top ranked retrieved images by the descriptors are not relevant to the queried image.

To compute the ANMRR the images in the retrieved set are assigned weights. If the low ranked images of the retrieved set are of the same class then they are assigned lower weights and the images with the higher rank of the retrieved set that belong to the same class as that of the queried image are assigned larger weights. Retrieved images that do not belong to the same class are assigned largest possible constant weights. Hence, the descriptors, which retrieve more number of low ranked relevant images that belong to the same class as that of the queried image have lower ANMRR. It can be seen from the experimental results that the proposed descriptor has lower ANMRR as compared to state of the art facial image descriptors.

Performance of the proposed method has been analyzed on the latest and most challenging facial image databases namely: Caltech-Face [29], CASIA-Face-V5-Cropped [30], Color FERET [31-32], and LFW [33]. Lengths of the proposed and other state of the art descriptors are shown in Table 1. Length of the proposed descriptor is half of the length of LDP and LVP. Three performance measures namely ARP, ARR, and ANMRR are used in all the experiments. ANMRR is computed for the maximum number of images in a particular category of the database.

Table 1: Length of the descriptors

| Descriptor (year) | Length (bits) | Length (bins) |
|---|---|---|
| CSLBP (2009) | 4 | 16 |
| CSLTP (2010) | - | 9 |
| LDGP (2015) | 6 | 64 |
| LBP (1996) | 8 | 256 |
| SLBP (2015) | 8 | 256 |
| LDP (2010) | 8×4 | 256×4 |
| LVP (2014) | 8×4 | 256×4 |
| LQPAT (Proposed) | 8×2 | 256×2 |

*4.1 Performance analysis on Caltech-Face database*

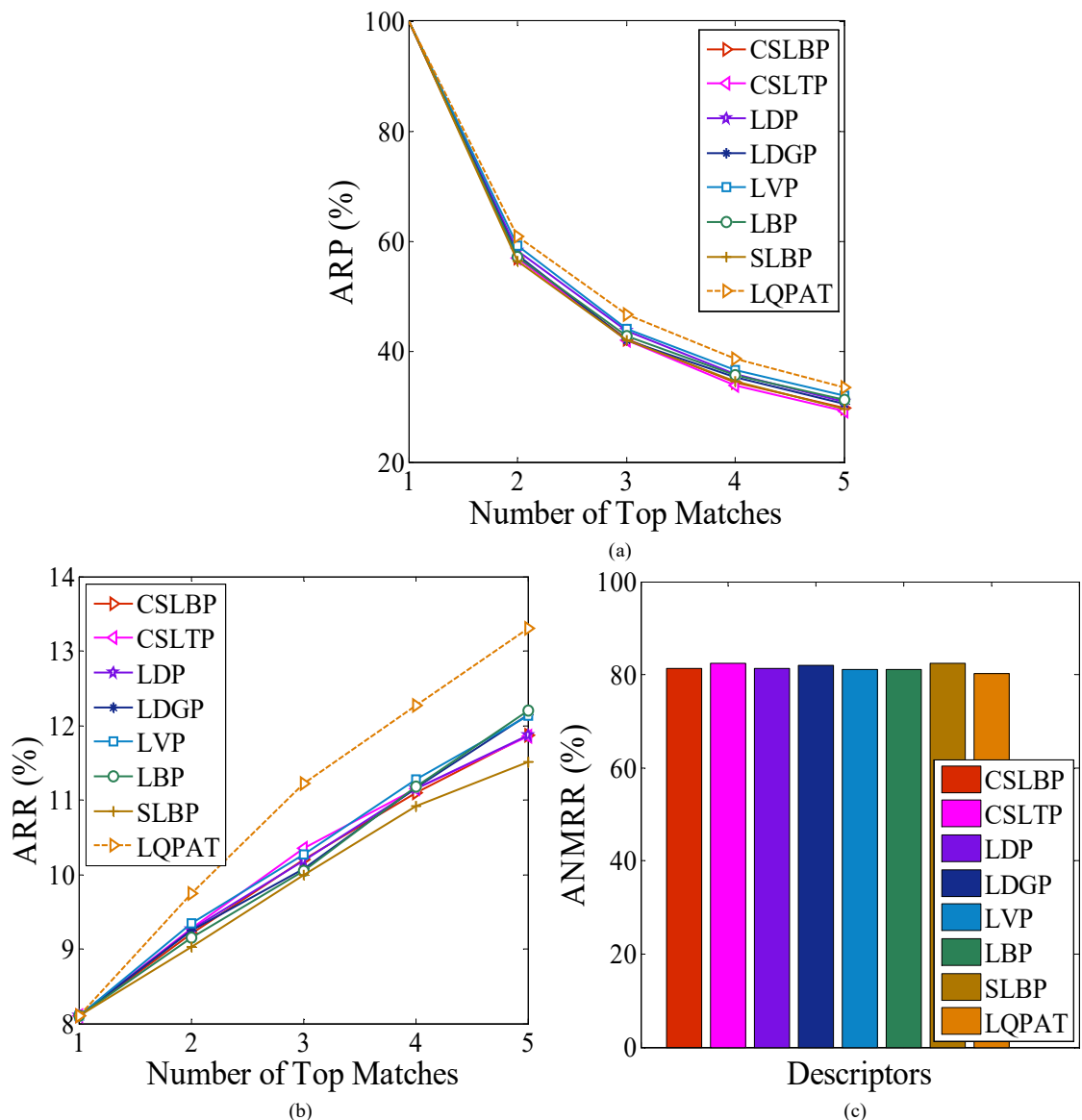

Fig. 5. (a) ARP, (b) ARR, and (c) ANMRR. computed over Caltech-Face database.

Caltech-Face database contains 450 frontal faces of 27 subjects with different backgrounds, expressions, and illuminations [29].

The ARP, ARR and ANMRR values computed over Caltech-Face database has been shown in Fig.5(a-c). The proposed method shows 2%-3% improvement in ARP over its nearest counterpart LVP. LQPAT achieves at least 0.5%-1% improvement over state of the art descriptors in ARR. Hence it can be concluded that LQPAT retrieves more relevant facial images as compared to its counterparts. A reduction of 1% in ANMRR achieved by LQPAT shows that it retrieves more images having lower rank (images with low rank are closer to queried image).

*4.2 Performance analysis on CASIA-Face-V5-Cropped database*

"Portions of the research in this paper use the CASIA-FaceV5 collected by the Chinese Academy of Sciences' Institute of Automation (CASIA)" [30]. CASIA-Face-5.0 database contains 5 color images each of the 500 individuals. Images are captured with intra-class variations such as illumination, pose, expressions, eye-glasses, and imaging distance [30].

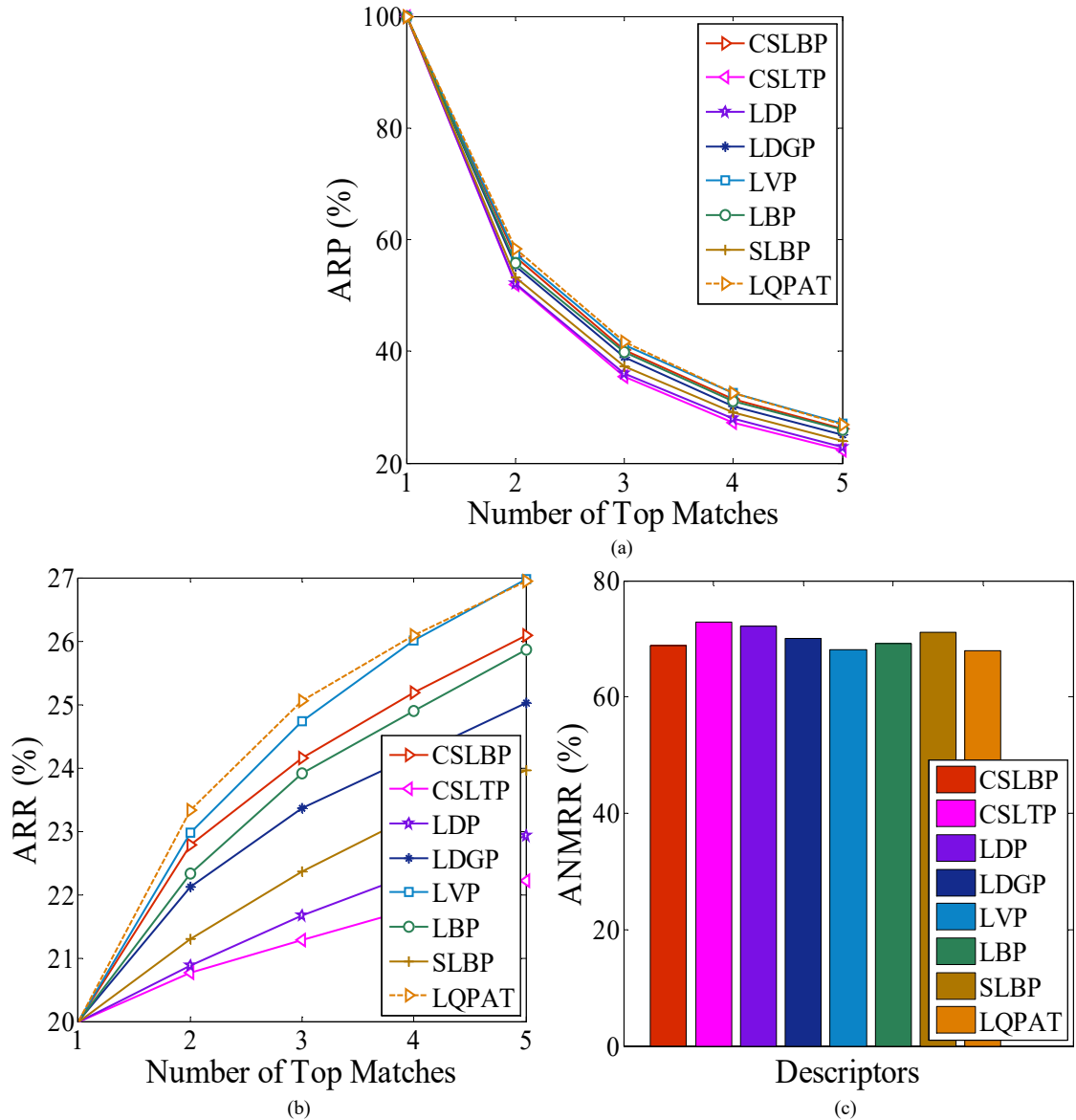

Fig. 6. (a) ARP, (b) ARR, and (c) ANMRR. computed over CASIA-Face database.

Proposed descriptor LQPAT achieves better ARP and ARR over CASIA-Face-5.0 database which is confirmed through the retrieval results shown in Fig. 6(a) and (b) than most recent descriptors such as LDP, LVP, and LDGP. LQPAT achieves approximately 2% better ARP and 1% better ARR over LVP, 4% better ARP and 1.2% better ARR over LDGP. As shown in Fig. 6(c) LQPAT achieves lowest ANMRR, which indicates that the most of the retrieved images are having lower/better rank or non-matching images are having higher rank.

*4.3 Performance analysis on Color FERET database*

"Portions of the research in this paper use the FERET database of facial images collected under the FERET program, sponsored by the DOD Counterdrug Technology Development Program Office". Color-FERET database is one of the most challenging facial image databases with severe variations in pose and expression. The color FERET database contains 11,338 facial images of 994 individuals at different orientations. There are 13 different poses used in the images of the database [31-32].

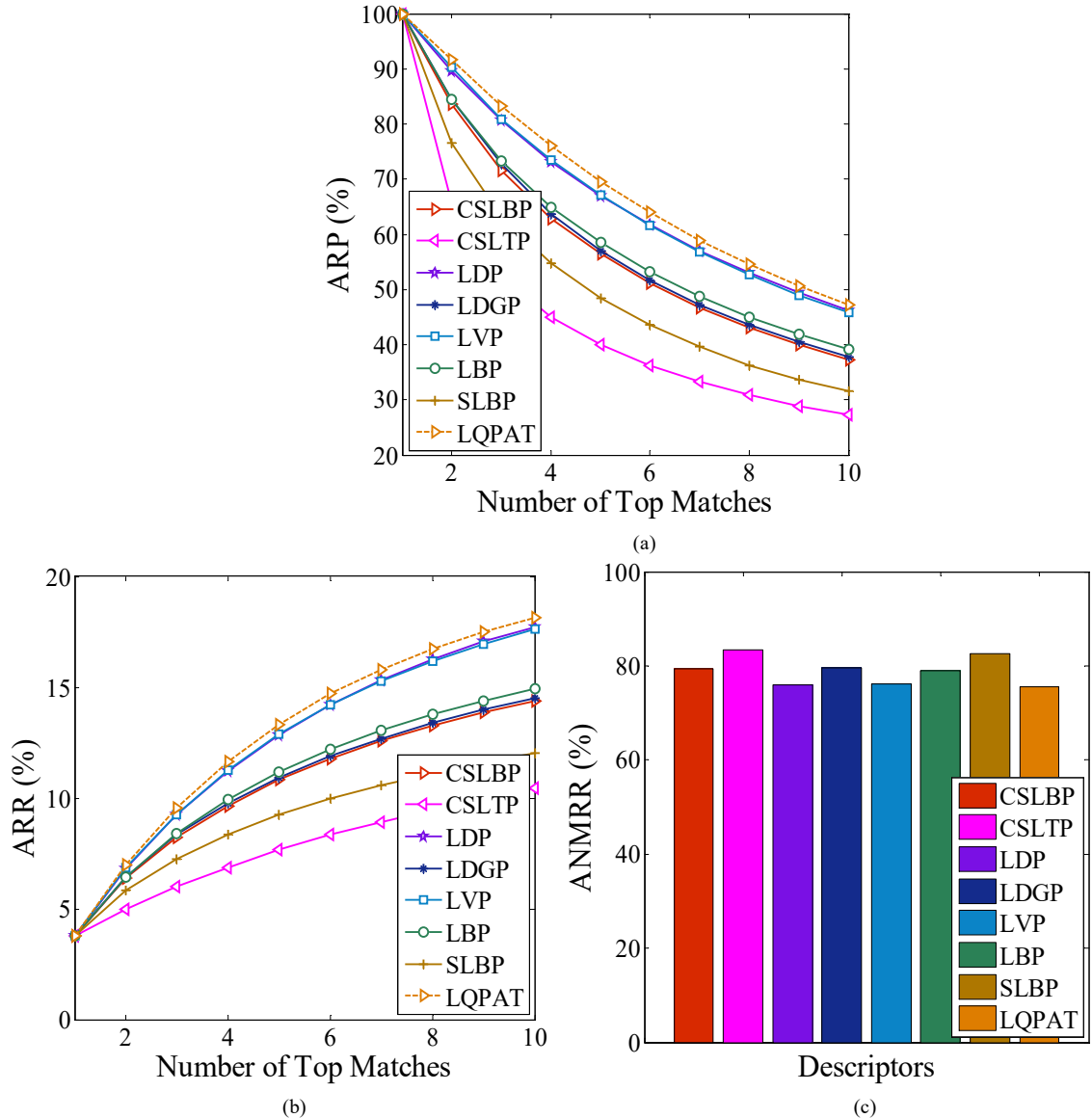

Fig. 7. (a) ARP, (b) ARR, and (c) ANMRR. computed over Color-FERET database.

The proposed descriptor is examined on color FERET database to show its capabilities against pose and expression variations. ARP and ARR values are shown in Fig. 7(a-b) for a maximum of 10 retrieved images. The ARP and ARR of LQPAT have gained approximately 2% of improvement over its nearest counterpart LVP, whereas, the dimension of LQPAT is approximately 0.5 times of LVP.

ANMRR of the proposed descriptor depicted in Fig. 7(c) is 1% lower than its nearest counterparts such as LVP, LDP, etc. It signifies that the proposed descriptor retrieves more relevant images with lowest possible ranks (images with low rank are closer to queried image).

### 4.4 Performance analysis on LFW database

There are 13,233 color facial images of 5,749 individuals in LFW database [33]. 1680 individuals have two or more images and rest of the individuals have only one image. As the proposed descriptor has been designed to work under unconstrained environment, we test the proposed descriptors over this database.

The descriptor is tested on LFW database as it is one of the largest and most challenging database of facial images taken under uncontrolled real world environment. Performance of the proposed descriptor has been evaluated on LFW to show its robustness against unconstrained variations in pose, illumination, and expression.

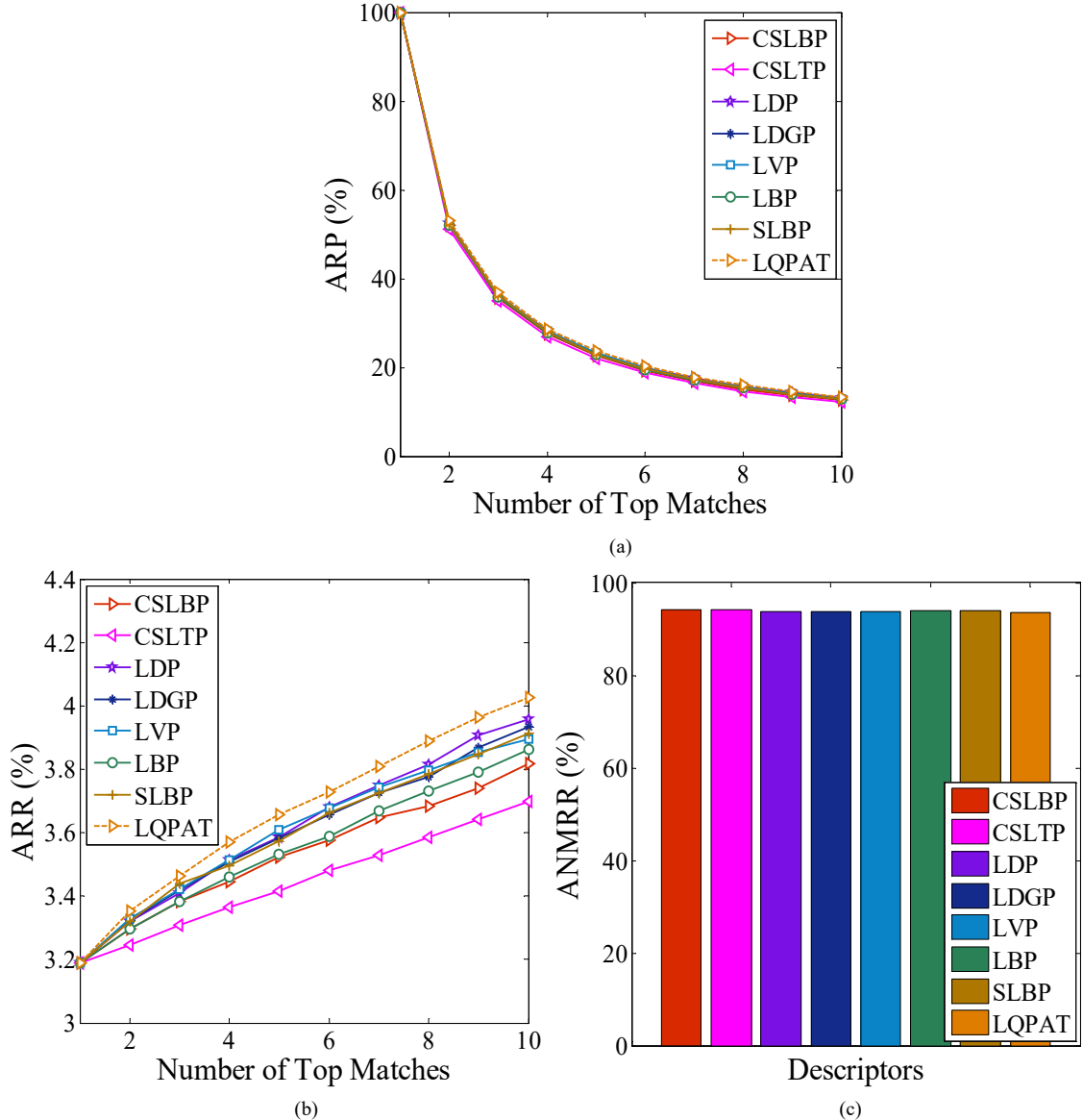

Fig. 8. (a) ARP, (b) ARR, and (c) ANMRR. computed over LFW database.

ARP and ARR values are depicted in Fig. 8(a) and (b) respectively. Improvements in ARP and ARR illustrate that LQPAT retrieves more relevant facial images as compared to other state of the art descriptors. Lowest value of ANMRR shown in Fig. 8(c) for LQPAT, which is at least 0.1% less compared to state of the art descriptors, indicates that it retrieves most of the low ranked images (low rank images are closer to queried image).

*4.5 Performance analysis in recognition framework*

Recognition rates are computed by taking each image in the database as probe and rest of the images as gallery. If there are $N$ images in the database then each image is taken as probe in turn and rest $(N-1)$ images are taken as gallery. The distance between probe feature and gallery feature is computed using $\chi^2$ distance. There are $(N-1)$ distances for each probe. The gallery image with lowest distance is given the lowest rank. If the gallery image with lowest rank belongs to the same class as the class of the probe image then it is taken as a match. The recognition rate of a descriptor is computed as

$$Recognition\ Rate = (number\ of\ matches/N)*100 \qquad (13)$$

The Cumulative Match Curves (CMCs) for different data bases are shown in Fig. 9. For CMC a match is taken, if the class of the probe image matches with the class of at least one gallery image with rank less than or equal to the maximum rank specified.

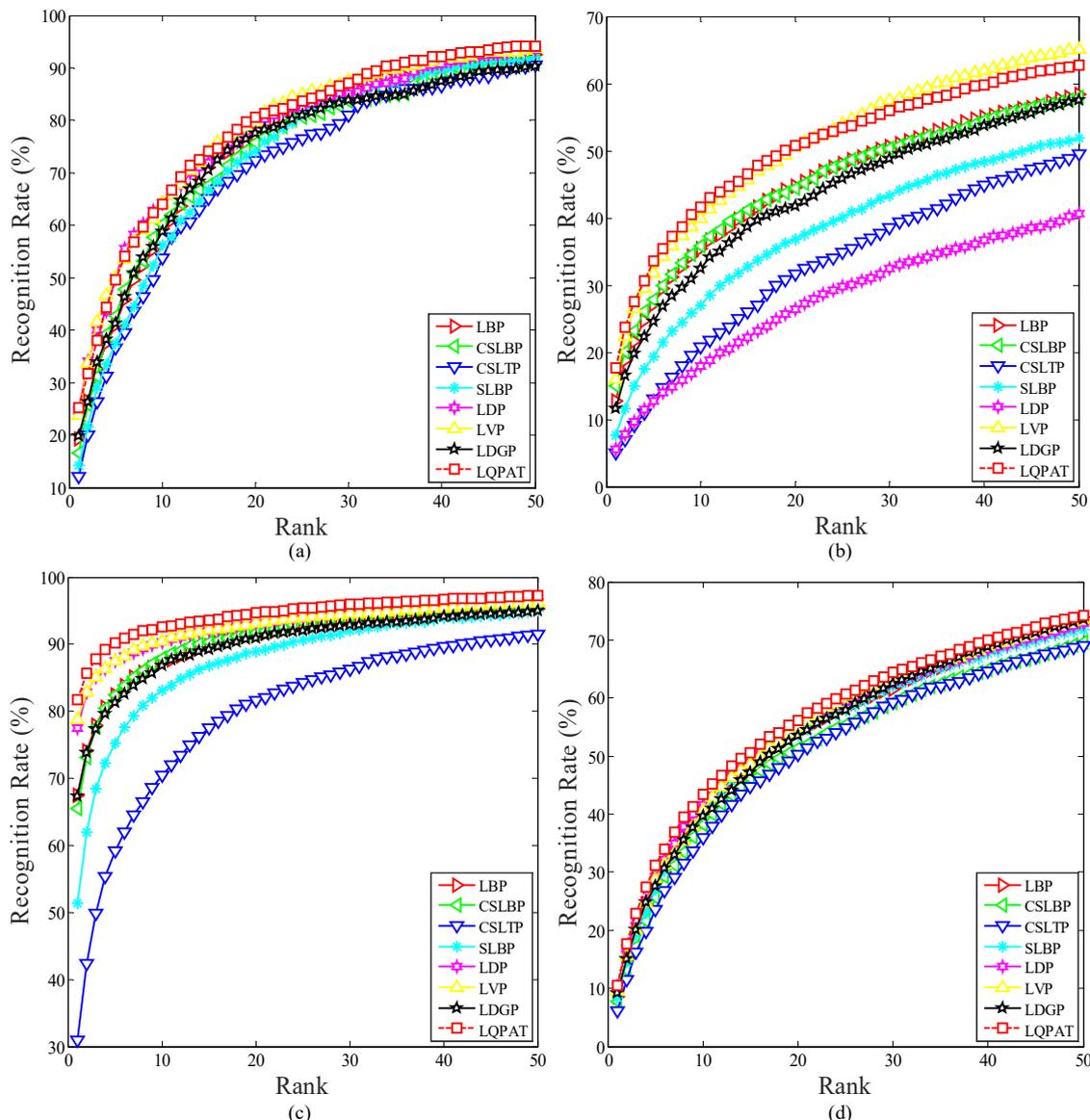

Fig. 9. CMC for different descriptors and LQPAT on databases (a) Caltech-Face, (b) CASIA-Face-V5-Cropped, (c) Color-FERET, and (d) LFW.

The proposed method shows consistent improvement in recognition rate with increasing ranks. There is a 1% improvement in recognition rate achieved by the proposed descriptor over its nearest counterpart LVP on Caltech_Face database. LQPAT shows 2% and 3% improvement over its nearest counterpart on Color-FERET and LFW databases respectively. The proposed method shows significant improvement of 3% over its nearest counterpart on CASIA-Face-V5-Cropped database. Significant and steady improvement shown by the proposed method on the most challenging databases illustrates the robustness of the proposed descriptor against pose, illumination, background and expression variations.

Average recognition rates shown in Fig.10 are computed by randomly dividing the dataset into disjoint probe and gallery sets of different size. Probe sets are prepared by randomly selecting 20%, 30%, 40%, 50% and 60% images from the datasets and the remaining images are used as corresponding gallery. 10 fold cross validation is used to calculate the average recognition rates for each probe and gallery pair the datasets. Average recognition rate is the average of recognition rates obtained in 10 iterations of the experiment for a particular probe and gallery pair.

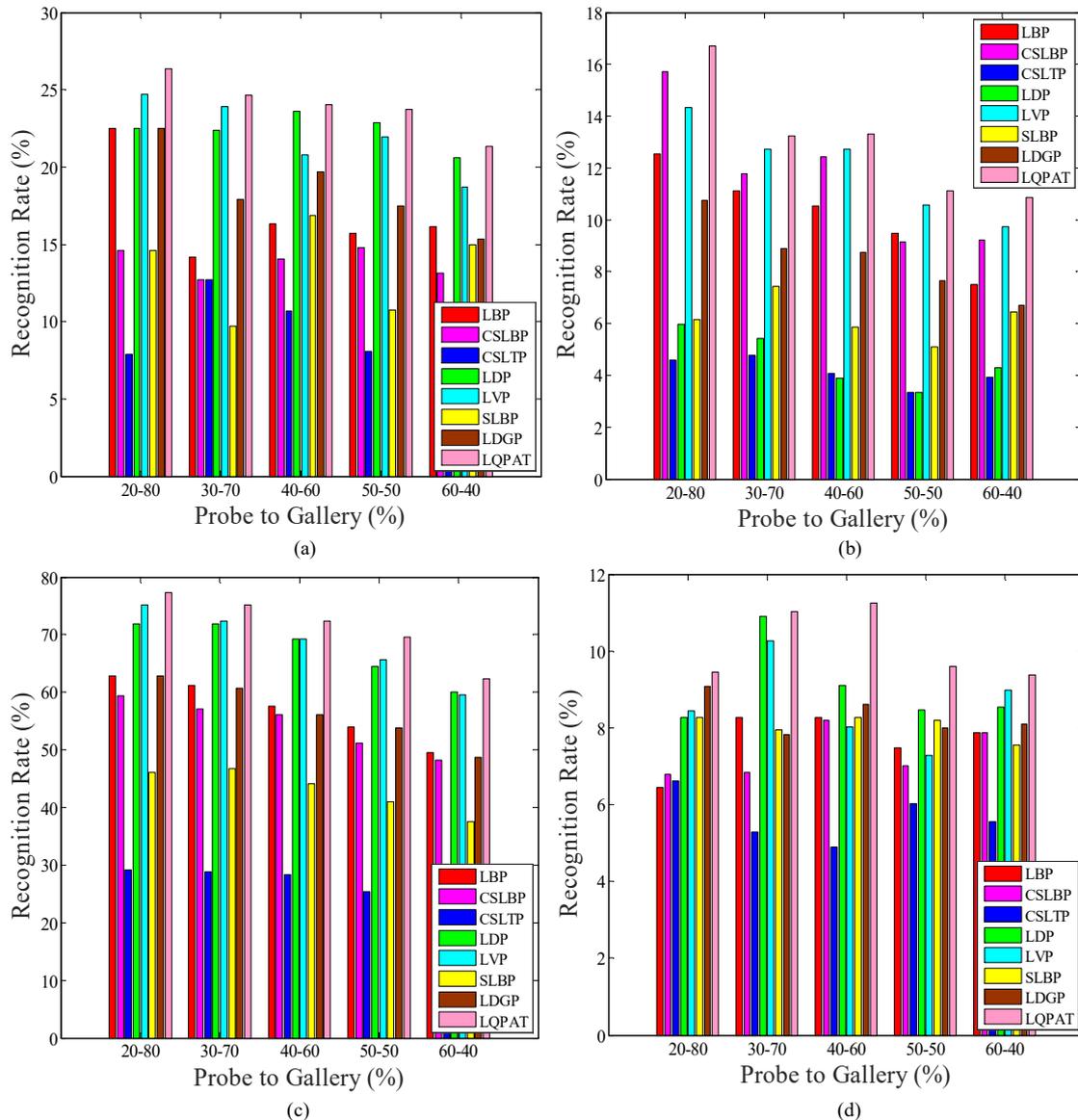

Fig. 10. Comparative average recognition rates of different descriptors and LQPAT of 10-fold cross validation with different sized probe and gallery on databases (a) Caltech-Face, (b) CASIA-Face-V5-Cropped, (c) Color-FERET, and (d) LFW.

LQPAT shows significant improvement over other descriptors on Caltech-Face database. Average recognition rates of LBP, CSLBP, CSLTP, LDP, LVP, SLBP, LDGP and LQPAT for 20%-80% ratio of the probe and gallery set are 22.47%, 14.60%, 7.86%, 22.47%, 24.71%, 14.60%, 22.47%, and 26.34% respectively. Average recognition rates of LBP, CSLBP, CSLTP, LDP, LVP, SLBP, LDGP and LQPAT computed over CASIA-Face-V5-Cropped database are 12.52%, 15.71%, 4.57%, 5.96%, 14.31%, 6.16%, 10.74%, and 16.71% respectively. The average recognition rates of LBP, CSLBP, CSLTP, LDP, LVP, SLBP, LDGP and LQPAT computed over Color-FERET database are 62.89%, 59.31%, 29.10%, 71.89%, 75.09%, 46.12%, 62.89%, and 77.31% respectively. Similarly average recognition rates of LBP, CSLBP, CSLTP, LDP, LVP, SLBP, LDGP and LQPAT computed over LFW database 40%-60% ratio of the probe and gallery set are 8.27%, 8.18%, 4.88%, 9.09%, 8.02%, 8.27%, 8.60%, and 11.24% respectively.

The performance analysis conducted in recognition framework shows that the proposed descriptor achieves consistent improvement over other state of the art descriptors with varying size of gallery. It also shows that the proposed descriptor attains better recognition rates with increasing size of gallery even on most challenging databases.

*4.6 Complexity Analysis*

The computational complexity of the proposed descriptor is comparable with the complexity of the state of the art descriptors LBP, SLBP, LDP, LVP etc. In the local neighborhood the proposed descriptor requires four comparisons to compute a four bit pattern. There are four such patterns which require 16 comparisons. Hence the proposed descriptor takes 16 comparisons to compute the micropattern in the local neighborhood of the reference pixel. The decimal conversions as shown in (4)-(7) require 12 additions and 16 multiplications which add up to 28 fundamental operations. Hence the total number of operations required to compute the feature image from an image of size $M \times N$ is $44 \times M \times N$. Therefore the computational complexity of the descriptor is $O(M \times N)$ which is computationally comparable with the complexity of the state of the art descriptors.

## 5 Conclusion

One of the major drawbacks of the existing descriptors was the consideration of only limited number of local neighbors. There are descriptors which tries to accommodate more neighborhood pixels. However length of such descriptors are very large. LQPAT is a local descriptor which captures distinctive relationships that exists amongst most of the pixels in the local neighborhood by expanding the neighborhood to four (quadruple) squares. The proposed descriptor accommodates 16 pixels of the local neighborhood, which increases the region of interest and decreases the intra-class dissimilarity. Length of the proposed descriptor is $2 \times 256$, which is half the length of LDP and LVP. There are descriptors with lesser feature lengths such as LBP, CSLBP, CSLTP, SLBP, LDGP. However, LQPAT outperforms these descriptors with respect to retrieval and recognition rates. The feature length can be further reduced with appropriate dimensionality reduction algorithms without affecting the accuracy. As the test results show, the proposed descriptor achieves better retrieval as well as recognition rates than the state of the descriptors under constrained as well as unconstrained environments.